# Scaling Construction Grammar up to Production Systems: the Situated Constructional Interpretation Model

Guillaume Pitel

Langue et Dialogue

LORIA

BP239 54000 Nancy, France

Guillaume.Pitel@gmail.com

**Abstract**

While a great effort has concerned the development of fully integrated modular understanding systems, few researches have focused on the problem of unifying existing linguistic formalisms with cognitive processing models. The Situated Constructional Interpretation Model is one of these attempts. In this model, the notion of "construction" has been adapted in order to be able to mimic the behavior of Production Systems. The Construction Grammar approach establishes a model of the relations between linguistic forms and meaning, by the mean of constructions. The latter can be considered as pairings from a topologically structured space to an unstructured space, in some way a special kind of production rules.

## 1 Introduction

Accounting for pragmatical and cognitive phenomena in a linguistic formalism is a challenging task whose resolution would be of great benefit for many fields of linguistics, especially those dealing with interpretation in a context. In domains such as practical dialogue or embodied understanding, there would be a real gain in dealing with environment data the same way one deals with linguistic data. These kinds of systems currently need *ad hoc* heuristics or representations. These heuristics are implemented in modules that are often impossible to reuse for another task than the one they were developed for. This point particularly concerns phenomena that lay at the interface of linguistics and general cognition, such as vagueness (Ballweg, 1983), reference resolution (Brown-Schmidt, 2003; Reboul, 1999), or modeling of cognitive representations (Langacker, 1983; Talmy, 1988).

Similarly, accounting for linguistic phenomena in a psychologically motivated model is far from simple. The attempts in that direction are often limited to simple phenomena, because all linguistic formalisms rely on principles slightly or totally different from those of cognitive architectures.

The definitive solution to this problem is probably still far from reach, but nevertheless, I think that the maturity of cognitive linguistics and the consequent emergence of language analyzers connected to cognitive architectures is an excellent direction toward a unified theory mixing linguistic and psychological models. The Embodied Construction Grammar or ECG (Bergen, 2003) and its analyzer (Bryant, 2003) are a good example of such an effort, even though it does not go beyond the linguistic layer since mental simulation is left to a mental simulation module based on the notion of x-schema (Narayanan, 2001).

Consequently, I try to propose a model that conciliates a linguistic theory with a cognitive architecture. The choice of the linguistic theory naturally goes to Construction Grammar (Fillmore, 1988; Kay 2002) and Frame Semantics (Fillmore, 1982), due to the parallel one can draw between a production rule and a construction, and the cogni-tive architecture is, obviously, the family of Production Systems (Newell, 1990; Anderson, 1993). Moreover, since many pragmatical models rely on topologically structured representation, I introduce the notion of **context**, a notion that has never been adapted to these theories in order to organize data in "storages" structured in dissimilar ways.

## 1.1 Typical Problem

Consider a situation where a user can command a software to manipulate some very simple objects (colored geometrical objects of various sizes). The user may say (a) "Put the small red square on the left", (b) "Remove the small red square on the left" or (c) "Move the small red square on the left".

First, these three utterances may involve different parsing depending on the actual environment of the utterance, at least for those with "put" and "move". Second, the "square" targeted by the user may be a rectangle in the actual software representation, with slightly different width and height. It may also be relatively small compared to other red squares, but bigger than other objects, and relatively red compared to other non-square objects.

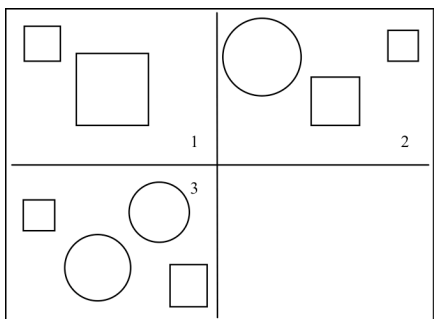

Figure 1: Some situations involving different understandings (without color).

Imagine what happens in the different situations illustrated in Figure 1. In situation 1, for instance, (a) would not be understandable, since the small square is already on the left, while (c) could lead to the one argument sense of "move", i.e. "move *something* somewhere else", not to the two arguments version "move *something somewhere*" (actually, the one-argument sense is an implicit understanding of the destination allowed by "move", so the difference should not be lexicalized). In situation 2, (b) and (c) would lead to two different interpretations of the referring expression "the small red square on the left": in (b), it refers to the square in the center (with a possible wavering), while it preferably refers to the square on the right in (c). In situation 3, (c) may be interpreted with the one argument sense of "move", and will target the square on the left since it is the smallest, but there should be a strong hesitation, since the other square is not that bigger, and the two arguments sense of "move" is intuitively preferred. At the same time (a) will target the square on the right, which is relatively small compared to the neighboring circles, but would raise incomprehension if the circles were missing.

In general, in order to take those facts into account, it is necessary either to produce all possible analyzes at each layer of the interpretation (which is quite problematic if it is desirable to allow for imperfect analyzes), or to allow two-ways interactions between the layers of interpretation (for instance, the pragmatic layer talking back to the semantic layer about the fact that the original position of an object is the same as the requested destination, which may indicate a wrong analysis).

My proposal is to allow for a generic capacity of interaction between the states of the interpretation (speaking about states is better than about layers since the latter presupposes something about the organizing of the interpretation), based on a unified operation between all the possible states. More specifically, the idea is to merge the notions from construction grammar and productions systems.

## 1.2 Merging Construction Grammar and Production Systems

Merging a linguistic analyzer with a cognitive processing model may seem a bit useless since they do not share the same objective. Linguistic analyzer's goal is to provide a formal model for the representation of linguistic knowledge, accordingly to linguistic observations. Cognitive models, on the other hand, aim at helping the modeling of real cognitive processing, in order to compare theoretical model of perception processing with real data from experiments. Cognitive models like production systems being Turing-equivalent, they typically do not lack of any expressiveness, meaning that anything one can describe with any linguistic representation could be implemented within a cognitive model (hopefully, since linguistic competence is part of the cognitive competence).

However, to my knowledge, no attempt to try describing a linguistic competence within a cognitive model has gone a long way. Existing researches on that topic have focused on very narrow problems, and what is more important, have been tightened to very small lexicons (Emond, 1997; Ball, 2003; Fowles-Winkler and Michaelis, 2005).

My analysis of this problem is that production systems are too permissive to allow a human to describe a grammar with a reasonable effort. More specifically, all generalization links that exist between grammar rules should be encoded in some explicit way in a production system.

Furthermore, linguistic formalisms are designed in such way to only express all possible human languages. In other words, a linguistic formalism is successful when it is flexible enough to describe all linguistic phenomena, while being human-readable enough to allow for a large-scale grammar development. As a consequence, linguistic formalisms are too restrictive to allow dealing with cognitive processes like the ones described using production systems.

Putting together a linguistic formalism and a model of pragmatical and cognitive processing implies to make a choice among all the current theories. Given the large predominance of production systems in cognitive modeling, it seems quit-natural to choose them as the cognitive model. The choice for the linguistic formalism is more open. Previous attempts of linguistic modeling in cognitive models have used $\overline{X}$ theory, categorical grammar or construction grammar. My pick has been the construction grammar because it shares some interesting features with production systems, and because it deals directly with semantic, contrary to other grammatical theories. Particularly, constructions are pairing between too poles: form and meaning, this is very similar to the notion of a production taking one input from a chunk, and producing its output into another one.

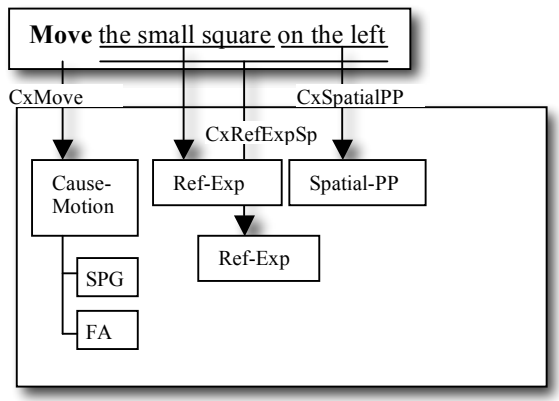

Figure 3: First step of analysis, the global construction (Imperative) is not active yet.

## 1.3 Example of processing

In such an approach, what should happen when interpreting "*move the small square on the left*" in situation 3 on the Figure 1? The first step of the analysis (simplified for sake of clarity), illustrated in Figure 3, shows how "move" produces a predicate that encompasses a *Cause-Motion* schema, itself evoking a *Source-Path-Goal* (SPG) and a *Force-Action* (FA) schema. The *CxMove* construction adds a constraint about the fact that *source* and *goal* should differ.

After this, two constructions *CxImperative* can connect, through their *theme* role, the referents evoked by the *RefExp* shemas (each construction being one possible interpretation) with the *source* of the *Source-Path-Goal*. The *CxImperative* encapsulates the predicate in a *Request* schema. Another construction can connect the *goal* of the *Source-Path-Goal* with the *Spatial-PP* produced from "on the left", with the predicate modified by the construction that took its *RefExp* from "the small square".

At this point, the "mental simulation" required

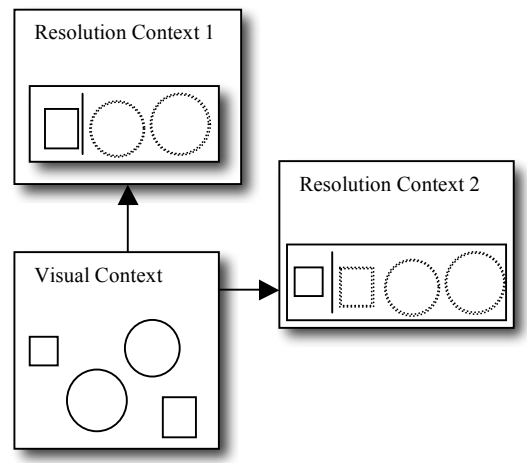

Figure 2: Mental simulation of the reference resolution

to resolve the referents can start. This step is illustrated in a very simplified way in Figure 2. The complete process is described in (Pitel, 2004; Pitel

& Sansonnet, 2003) and processes potential referents through several sorting steps, one for each referential predicate (here: *square* and *small* in Resolution Context 1 from the two-arguments move interpretation; *square*, *small* and *on the left* in Resolution Context 2 from the other one). The process is described with the kind of constructions defined by the SCIM.

## 2 Basic Notions of the SCIM

The Situated Constructional Interpretation model (SCIM) describes how information can be processed in a way that is both linguistically and psychologically plausible. It relies on three notions: **schemas** are for low-level data description, **contexts** are for describing the organization of instances of schemas, and **s-constructions** represent the mean to process data. Eventually, a SCIM-based interpretation system will run instances of s-constructions that take and produce instances of schemas situated in instances of contexts. These three notions are partly inherited from the ECG.

### 2.1 Schemas

Schemas are *constrained, typed features structures,* with an *inheritance* mechanism and no type disjunction. Schemas are a kind of data type. They describe complex structures of information used to represent the state of the running interpretation. As shown in Figure 4, schemas are defined with three blocks:

- **inherits** *schema-name$_1$, …* which specifies from which schema(s) this one inherits from. a specific case of the schema *x*, it inherits all of its properties (roles and constraints).

- **roles**, which specifies a list of roles, constrained to a given schema type or atomic type (Integer, Boolean, String, or user-defined enumerations of symbols).

- **constraints**, which specify the constraints that must be verified in order for an instance of the schema to be a valid one. A constraint can be a *predicate* if the role has an atomic type, or an *identification* constraint (asserting that two roles must share the same value), or a *filler* constraint with a constant value.

An **instance of schema** is moreover described by values attached to its roles (some or all of them may be left underspecified), a unique identifier, a positive value representing its informative capacity, a percentage of trust level, and the list of its parents' identifiers. A parent of an instance of schema is an instance of schema "used" in the process that led to its production. It is thus possible, in a s-construction, to know whether two given instances of schema are somehow related to each other in the interpretation process.

```
schema <schema-id>
inherits <schema-id₀, ..., schema-idₙ>
Roles
[?]<local-type-id>:<atomic-type-id>
[?]<local-context-id>:<context-id>[@<local-context-id>]
[?]<local-schema-id>:<schema-id>[@<local-context-id>]
Constraints
<boolean-operation>(<constraint₀>, ..., <constraintₙ>)
<role-id> ← <atomic-value>|<function>(<atomic-value>,…)
<role-id> ↔ <role-id>|<C-function>(<role-id>)
<role-id> = <role-id>
<boolean-predicate>(<role-id₀>, ..., <role-idₙ>)
```

a <*role-id*> is one of:
  self (optional if not used alone)
  <local-type-id>
  <local-context-id>
  <local-schema-id>
  <inherited-schema>*<inherited-role-id>
  <role-id>.<sub-role-id>

Figure 4: Schema definition formalism.

From the production systems perspective, schemas define the type of features that can be attached to a category. Basically, in that point of view, an instance of schema is a *chunk* and roles are *slots*.

### Schemas hierarchy

Schemas can inherit roles and constraints from other schemas. That means that schemas are organized in a multiple inheritance hierarchy. In order to avoid ambiguity in role access, inherited roles must be accessed through an inheritance path. For instance, accessing the role *color* in a schema *Square*, if the hierarchy is Figure→Rectangle→Square, and where the color role is declared in the Figure schema, would be realized through this kind of path: Rectangle*Figure*color.

Inheritance also means that an instance of schema S can be unified with a role whose type is R if S == R **or** if R is one of the parents of S.

One problem with this approach of inheritance is that, in order to fulfill the Liskov substitution principle (Liskov, 1988), it is sometimes necessary to use unnatural type hierarchies (stating that Square doesn't inherit from Rectangle, for instance). I am very mindful about this problem, since such a discrepancy is quite tedious for a model that aims to approximate the human way of processing information, but this problem is out of the scope of this paper[1].

**Constraints**

A schema declaration contains a set of constraints that must be satisfied in order for an instance of this schema to be considered valid. Constraints are specified with six basic forms:

- Type constraints on roles.

- Boolean operation (OR, NOT, NAND,…) connecting several constraints.

- **Filler** constraint symbolized by a single arrow (←) specifies that a constant, atomic value must fill the role in an instance.

- **Identification** constraint, symbolized by a double-headed arrow (↔), specifies that both sides of the constraint must unify, that is, all roles' values must be compatible with each other.

- An equality constraint (=) that constrains two roles to refer to the same instance.

- A **boolean predicate** constraint can be asserted between any number of roles.

Another kind of constraint, on the places occupied by instances of schema in context, will be explained in the section about s-constructions, as will the role of interrogation marks in the schema declaration formalism.

## 2.2 Contexts

A *context* declaration is a description of a container that can hold instances of schemas. In other words, it describes a space (including the topology part that can be specified by a set of relations and operations) that can contain pointers to instances of schemas at given *places*.

The notion of context inherits all of the properties of the notion of schema. Actually, a context is really a kind of schema and, as a consequence, a schema's role can be restricted to be a context. A declaration of context adds three more blocks to the declaration of a schema, as shown in Figure 5:

- **places** declare a list of *opaque* types (the internal structure of the type is hidden in the implementation) that describe an acceptable position in the context. Instances of schema (or context) that will be contained in an instance of this context will be linked with a position whose type is one (and only one) of the declared *places*. Examples of places are: *point, segment, multi-segment, line, box, disc, ...*

```
context <context-id>
inherits <context-id₀, ..., context-idₙ>
Roles
  // idem schemas roles
Constraints
  // idem schemas constraints
Places
  <place-id>
Relations
  <relation-id(<place-id>, <place-id>,...)> ↦ <type-id>
  // for instance: before(point, point) ↦ Boolean
Operations
  <operation-id>(<place-id>, <place-id>,...) ↦ <place-id>
  // for instance: intersection(segment, segment) ↦ segment
```

Figure 5: Context definition formalism

- **relations** are functions that associate a value in an atomic domain from one or more *places*. Relations define constraints on the positions of a set of instances of schema. For instance, one can define a precedence relation in a linear context.

---
[1] We consider that this problem could be solved by the approach called "Points of View Theory" (which is not related to inter-person points of view), proposed by Pitel (2004). In this theory, there is no type hierarchy, and the ability to substitute a representation by another is described by rules that can take the dynamic context into account. In this approach, types do not represent concepts, but points of view on perceptions (in the wide meaning), and transition from one point of view to the other is context-dependent.

- **operations** are functions that associate a *position* from one or more *positions*. For instance, a union of segments is an operation.

Terminologically, an instance of schema (or context) located in a context, that is, an instance with a place, will be called a **situated instance**, whereas an instance of schema (or context) simply connected to another instance by a role will just be called a **role instance**.

The only explicit equivalent to contexts in ECG is the notion of **space**, which describes Fauconnier's mental spaces (Fauconnier, 1985). Implicit contexts are however used in Construction Grammar: the **form pole**, which stores instances of schemas representing linguistic data in a linear space, and the unstructured **meaning pole**.

---

s-construction <s-construction-id>
**inherits** <s-construction-id$_0$, ..., s-construction-id$_n$>
**roles** // idem schema's roles
**constructional**
 <local-s-constr-id>: <s-construction-id>
**constituents**
 <local-ctx-id>: <context-id>[@<local-ctx-id>]/I|O|IO
 <local-constit-id>: <schema-id>[@<local-ctx-id>]/I|O|IO
**constraints**
 // idem schemas constraints, plus :
 // a role-id can be marked as muted: ?<role-id>
 // a place-id is either a <local-constit-id> or the result of a context operation like:
 // <local-ctx-id>.<context-operation-id>(<place-id>, ...)
 <role-id> ⊂ <role-id> // right hand side must be parent
 <local-ctx-id>.<context-relation-id>(<place-id>, ...)
 OUT(<local-constit-id>) // remove the situated instance

---

Figure 6: S-construction definition formalism

### 2.3 S-constructions

S-constructions are situated constructions, that is, constructions that describe the relations between several instances of schemas located in structured contexts. As for the notion of context, the notion of s-construction is derived from the schemas, because the s-construction itself can hold information. Besides that, the declaration of a s-construction contains:

- A **constructional** block that describes the other instances of s-constructions this s-construction relies on. The block contains a list of label: s-construction-name declarations. Any restriction on the constituents of those instances of s-construction is described as a constraint on label.constituent in the **constraints** block.

- A **constituents** block that describes the instances of contexts and schemas constrained by the s-construction (note that the meaning of constituents is different than in ECG). The declaration of those constituents specifies whether the instance must preexist and/or whether it may be created or specified by the s-construction's constraints. From a production system point of view, it means that we describe which instances are in the input, and which one are produced.

### S-constructions hierarchy

Like schemas, s-constructions are organized in a multiple inheritance hierarchy. Moreover, s-constructions benefit from a mechanism of *constructional dependence*, held by the constructional block. Those two notions are, to some extent, redundant. Indeed, inheriting from a s-construction is equivalent to having an instance of this s-construction in the constructional block. However, one can have two different instances of the same s-construction in the constructional block, whereas it is impossible to inherit twice from the same s-construction. Moreover, it is possible to add a negative semantics in the constructional block, in order to assert that some instance of s-construction must not have occurred to satisfy the s-construction's conditions.

The *constructional* block is thus more powerful than the classical inheritance relation, but as for the schemas hierarchy, it is not within the scope of this paper to discuss about the inheritance relations between s-constructions. A declaration of s-construction is thus, from that point of view, in conformance with the standard view shared in construction grammars.

### Situated aspects of s-constructions

A s-construction can "choose" instances of schemas, given positional constraints in the context where the instances of schemas are stored. Then, the s-construction will "create" new instances of context or schemas, or will specify some previously underspecified role's value. S-constructions can connect together more than two instances of schema. To that extent, it differs from ECG's construction (ECG's way of doing so makes use of an evoke block).

The specification of structural constraints is very similar to the other constraints. A structural constraint looks like this: context-id.relation(roles-in-context-id). Basically, a context relation is considered as a boolean predicate constraint. The main difference is that, instead of specifying the roles, such a constraint specifies the place of the instance of schema referred to by the role.

## Dynamic aspects

The biggest gap between productions systems and construction grammar is the difference between the dynamic nature of productions versus the declarative nature of linguistic constructions. For instance, a typical rule in a production system (from the ACT-R tutorial) would be represented in Figure 7. In order to take this possibility into account, it is necessary to introduce at some point some imperative features in the s-construction.

Imperative features are introduced through several mechanisms. The first one is about role instances, the second one is about situated instances and the third one is about specifying constituents acting as inputs and/or outputs.

| ACT-R declaration | English description |
|---|---|
| (p start | |
| =goal> | If the goal is |
| ISA count-from | to count from |
| start =num1 | the number =num1 |
| step start | and the step is start |
| ==> | Then |
| =goal> | |
| step counting | to note that one is now counting |
| +retrieval> | and request a retrieval |
| ISA count-order | of a count-order fact |
| first =num1 | for the number that follows =num1 |
| ) | |

Figure 7: Example of ACT-R rule with a value changing

- **Mutable roles**. In the **roles** blocks, they are specified by a question mark (?). If a role is marked as mutable in a schema declaration, then it can be accessed through two means in a s-construction constraint. The usual way constrains the state of the role instance *before* the application of the s-construction, the mutated way constrains the state of the role instance *after* the application of the s-construction.

- **Removable situated instances**. The constraint *OUT(<constituent-id>)* specifies that the situated instance must be marked as not being present anymore in its context, after execution of the s-construction.

- **Input and/or output constituents**. Each constituent of a s-construction is marked with a symbol /I or /O, stating whether the situated instance should be present before and whether it will be modified.

## 3 Computational Aspects

Given the characteristics of the SCIM, its expressiveness and its procedural orientation, one cannot occult the problems that it raises from the computational point of view. Building an implementation of the Situated Constructional Interpretation Model definitely means to give up the idea of conducting a complete exploration of the search space.

The main problem is that two s-constructions may lead to contradictory constraints. In other words, one must keep track of all the decisions and explore all the possibilities.

The problem is even worse with mutable instances, since some constraints may be satisfied at some moment in one possible interpretation, while being unsatisfied at another moment. This time dependence must be handled very carefully, and adds some complexity to the processing of constraints.

However, the model also presents some interesting features, computationally speaking. For instance, it is quite easy to add a weighting layer to the SCIM, in order to simulate expectation, informational potential, or execution cost. Such a layer could be trained to learn how to lead to the best interpretations at a minimal cost.

## 4 Conclusion

In this paper I propose and describe a model of interpretation both linguistically and psychologically motivated. This model allows describing a construction grammar as well as a production system, with three basic notions: *schemas*, *contexts* and *s-constructions*. Applications for such a model are wide, from more integrated dialogue systems to a unified theory of cognition and language.

A longer description of the processing architecture would be necessary in order to really confront the hypotheses I made in the section "Computational aspects", but nevertheless, one can already draw a parallel between this model with a spatial structuring of information, and the structure that neuromimetic models can handle. Also, incomplete exploration of the search space, guided by a cost/gain approach, has previously been proposed as a plausible model of processing for human cognition. More than computational efficiency, the goal of this model is to propose a formalism that would be easier to use both for linguistic and cognitive modeling, in order to observe and act on the simulated processing of language and other cognitive functions.

Many of the claims in this paper have yet to be proved through the implementation of the SCIM, and cognitive modeling using the system. Since many processing models have been made both on construction grammar and production systems, important researches should be easy enough to re-use in the SCIM.